\documentclass[a4paper,fleqn]{cas-dc}

\usepackage[numbers]{natbib}
\usepackage{multirow}
\usepackage{textcomp}
\usepackage{stfloats}
\usepackage{url}
\usepackage{verbatim}
\usepackage{graphicx}
\usepackage[normalem]{ulem}
\newcommand{\Methods}[2]{\begin{tabular}[c]{@{}c@{}}{#1}\\ {#2}\end{tabular}}

\def\tsc#1{\csdef{#1}{\textsc{\lowercase{#1}}\xspace}}
\tsc{WGM}
\tsc{QE}
\tsc{EP}
\tsc{PMS}
\tsc{BEC}
\tsc{DE}

\begin{document}
\let\WriteBookmarks\relax
\def\floatpagepagefraction{1}
\def\textpagefraction{.001}
\let\printorcid\relax
\shorttitle{TransPose: 6D Object Pose Estimation with Geometry-Aware Transformer}

\shortauthors{Xiao Lin et~al.}

\title [mode = title]{TransPose: 6D Object Pose Estimation with Geometry-Aware Transformer}                      
\tnotemark[1]

\tnotetext[1]{This paper is supported by the National Natural Science Foundation of China (No. 62073245). Shanghai Science and Technology Innovation Action Plan (22511104900).}


%
\author[1]{Xiao Lin}
\ead{2111118@tongji.edu.cn}
\cormark[1]



\credit{Conceptualization, Methodology, Coding, Writing original draft}

\affiliation[1]{organization={College of Electronics and Information Engineering, Tongji University},
    city={Shanghai},
    postcode={201804}, 
    country={China}}

\author[1]{Deming Wang}
\credit{Conceptualization, Methodology, Coding}
\cormark[1]

\author[1]{Guangliang Zhou}
\credit{Writing, reviewing, editing}

\author[1]{Chengju Liu}
\credit{Supervision, Writing, reviewing, editing}
\cormark[2]
\ead{liuchengju@tongji.edu.cn}
\author[1]{Qijun Chen}
\credit{Supervision, Writing, reviewing, editing}
\cormark[2]
\ead{qjchen@tongji.edu.cn}





\cortext[cor1]{Equal contribution}
\cortext[cor2]{Corresponding author: Chengju Liu, Qijun Chen}



\begin{abstract}
    Efficient and accurate estimation of objects' pose is essential in numerous practical applications. 
    Due to the depth data contains abundant geometric information, some existing methods devote to extract features from 3D point cloud. 
    However, these depth-based methods focus on extracting the point cloud local features and consider less about the global information.
    How to extract and utilize the local and global geometry features in depth information is crucial to achieve accurate predictions.
    To this end, we propose \textbf{TransPose}, a novel 6D pose framework that exploits Transformer Encoder with geometry-aware module to develop better learning of point cloud feature representations.
    To better extract local geometry features, we finely design the graph convolution network-based feature extractor that first uniformly sample point cloud and extract point pair features of point cloud.
    %
    To further improve robustness to occlusion, we adopt Transformer to perform the propagation of global information, making each local feature obtains global information.
    Moreover, we introduce geometry-aware module in Transformer Encoder, which to form an effective constrain for point cloud feature learning and makes the global information exchange more tightly coupled with point cloud tasks.
    Extensive experiments indicate the effectiveness of TransPose, our pose estimation pipeline achieves competitive results on three benchmark datasets. 
\end{abstract}




\maketitle
\section{Introduction}\label{sec:intro}
6D Pose Estimation is an important branch in the field of 3D object detection and plays a significant role in lots of real-world applications, such as augmented reality~\cite{marchand2015pose}, autonomous driving~\cite{chen2017multi} and robotic manipulation~\cite{tremblay2018deep}. 
The research focuses on rigid bodies with the aim of determining the transformation between the coordinate system of the target object relative to the coordinate system of the visual or laser sensor. It has been proven a challenging problem due to sensor noise, varying illumination and occlusion.

Recently, some researchers have applied deep neural networks to estimate 6D object pose from a single RGB image~\cite{peng2019pvnet,park2019pix2pose,yang2023er,JIANG202216} and achieved promising results. 
However, RGB-based methods are very susceptible to
illumination changes and occlusions, which limit the performance of these approaches in complicated scenarios.
What's more, the lack of depth information in RGB images prevents such methods from obtaining accurate 6D object pose.

\begin{figure}[htbp]
    \includegraphics[width=\columnwidth]{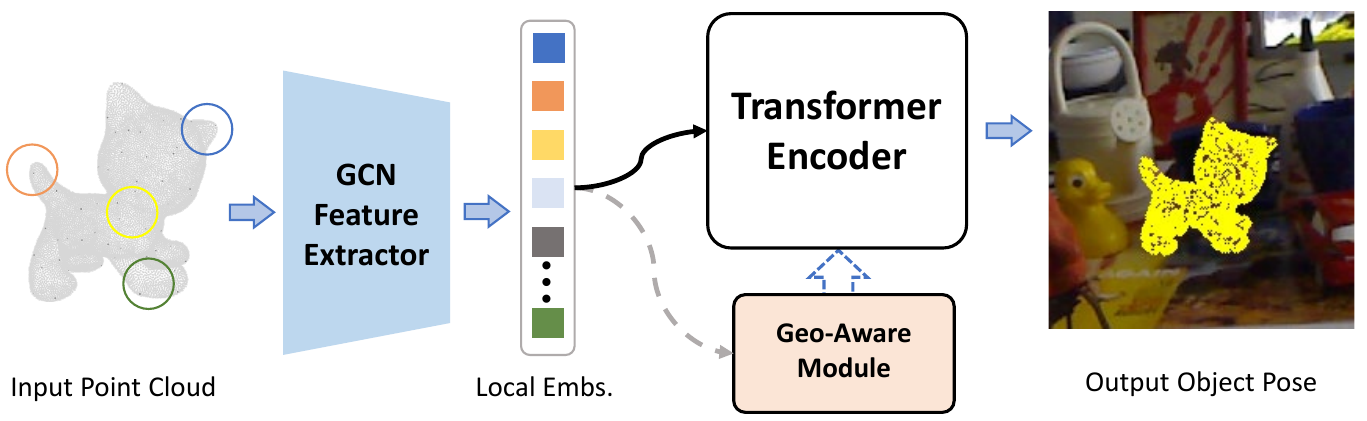}
    \caption{\textbf{Illustration of TransPose}. Given point cloud of objects as input, the model uniformly samples several local regions of the point cloud and extracts local neighborhood features via local feature extractor base on graph convolution network. 
    The obtained feature form a point cloud embeddings, which is fed to a transformer encoder with the geometry-aware module to obtain the global features. Finally, the pose estimation network recovers object 6D pose parameters.
    }\label{fig:framework_simple}
\end{figure}

Compared with RGB images, point clouds can provide a wealth of spatial geometry structure information and topological relations of the point cloud. Naturally, methods based on point cloud are more appropriate in complicated scenarios. However, it is quite challenging to process point clouds using convolution neural networks like 2D vision tasks due to the irregularity of point clouds. 
How to obtain the geometric features of objects more effectively is the key challenge to point clouds-based object pose estimation methods.
The PointNet series~\cite{qi2017pointnet,qi2017pointnet++} is the pioneering effort that applies Multi-Layer Perceptions(MLPs) to process original point clouds directly. Furthermore, they devise hierarchical structures to learn local high dimensional features with increasing contextual scales. 
Essentially, PointNet series migrates 2D CNN to 3D point cloud to learn the spatial encoding of each individual point features and then aggregate single point to a global point cloud signature. 
Though effective, these methods suffer from information loss during the process of downsampling and pooling. 
To better extract the spatial information, several works attempt to model point cloud as graph structure to obtain spatial features. 
GNN6D~\cite{yin2021graph} performs graph convolution operation to learn inner spatial information of point cloud and fuse the appearance feature with geometry feature. DGECN~\cite{cao2022dgecn} leverages local graph and edge convolution to assist in establishing 2D-3D correspondences.
Nevertheless, these works only obtain local information and consider less about the global propagation and exchange of information. Hence, there is still room for improvement in complicated scenarios. 

More recently, Transformer~\cite{vaswani2017attention} is introduced to the computer vision tasks and achieves remarkable results~\cite{devlin2018bert,dosovitskiy2020image}, which lead some researchers exploit it to capture better global feature representations of point clouds~\cite{guo2021pct,zhao2021point}. 
Transformer is an attention-based framework, which is first proposed in the field of natural language processing (NLP), and it has been proven to be efficient for the tasks involving long sequences due to the self-attention mechanism. 
However, Transformer has no inherent inductive bias for 3D visual tasks, which refers to a set of prior beliefs and assumptions that guide the learning process and assist the algorithm to make better predictions based on the available data.
The inductive bias plays the role of an inherent constraint in traditional visual models. 
For instance, CNN assume that pixels in the same region will have similar features while RNN views that the current state is only dependent on the previous states and is independent of time.
Existing works on point clouds learning based on Transformer attempt to design input point cloud sequences to be more appropriate for the encoder~\cite{zhao2021point} or introduce other attention mechanisms like cross-attention~\cite{guo2021pct,pan20213d}, but the lack of inductive bias in Transformer has not been fully investigated.

To this end, we propose a novel 6D pose estimation framework that adopt \textbf{Trans}former Encoder with geometry-aware module to fulfill 6D object \textbf{Pose} (TransPose) estimation task.
Our framework utilizes only the depth information as input to estimate the 6D pose of the object, as shown in Figure~\ref{fig:framework_simple}. The key insight of TransPose is that geometry and topology relations in point cloud can provide a guidance for the exchange of global information. 
Specifically, we first uniformly sample the point cloud into several local regions. To fully extract the local features, we finely design a novel local feature extractor base on graph convolution network (\textbf{GCN}) thanks to the great representations power of graph structure for topology information.
However, it is hard for local features to tackle the complicated scenarios like occlusion. We require local features to contain global information. Thus, we exploit strong associative representational capabilities of Transformer to achieve global information exchange. 
Furthermore, we introduce a geometry-aware module as inductive bias to form an effective constraint for feature learning of Transformer Encoder, making the global information exchange tightly coupled with the point cloud task.
Ablation studies have been performed to validate the effectiveness of the geometry-aware module, and we also conduct experiments on three popular benchmark datasets to fully evaluate our method: LineMod, Occlusion LineMod and YCB-Video datasets. Experimental results show that the proposed approach achieves impressive performance while employing only point cloud and is comparable to the state-of-the-art methods using RGB-D images.

In summary, the main contributions of this work are as follow:

\begin{itemize}
    \item We propose a novel 6D pose estimation framework that allows geometry relations of point cloud provide the guidance for exchange of global information.
    \item We finely design graph convolution network for local point cloud feature extraction and geometry-aware module to provide effective constraints for the Transformer.
    \item We demonstrate that our method can effective learn local and global spatial information from point cloud. We achieve competitive results on the LineMod, Occlusion LineMod and YCB-Video datasets.

\end{itemize}

The rest of the paper is organized as follows. Section \ref{sec:RelatedWorks} reviews several previous works on object pose estimation, graph convolution network and vision Transformer. The geometry-aware Transformer and proposed object pose estimation pipeline are detailed in Section \ref{sec:Method}. Furthermore, we report and analyze the experimental results in Section \ref{sec:Experiments}. Finally, we discuss the conclusion and future work in Section \ref{sec:Conclusion}.

\section{Related Work}
\label{sec:RelatedWorks}
\subsection{6D Object Pose Estimation}
\textbf{Pose Estimation with RGB Data.} 
One line of methods seek to establish a sparse or dense 2D-3D correspondence, and then apply Perspective-n-Point (PnP) to calculate the 6D pose. CDPN~\cite{li2019cdpn} propose to disentangles the pose to predict rotation and translation separately. 
DPOD~\cite{zakharov2019dpod} divides the continuous coordinate space into discrete space and classifies each pixel of 2D object surface. 
ER-Pose~\cite{yang2023er} predicts the direction and distance to a certain object keypoint from all object pixels within the range of object edge representation.
The other line of methods predict the parametric representation of the 6D pose of objects directly by means of deep neural networks, typically modeling the pose estimation task as a regression or classification task. 
PoseNet~\cite{qi2017pointnet} introduces the GoogleNet framework to perform camera relocalization directly via single RGB image. PoseCNN~\cite{xiang2017posecnn} designs two independent branches to estimate 3D position and 3D rotation respectively. 
MLFNet~\cite{JIANG202216} proposes the surface normals in the object coordinate system as an intermediate representation of pose
However, the loss of geometry information due to perspective projections limit the performance of these RGB only methods.

\textbf{Pose Estimation with depth Data.} With the dramatic development of depth sensor and the point cloud learning techniques, several depth data only methods gradually emerge. 
Naturally, the geometry information embedded in the depth data is more suitable for weak-texture scenarios. 
Wen et al.~\cite{wen2020robust} presents a depth-based framework to detect the adaptive hand’s state via efficient parallel search.
G2L-Net~\cite{chen2020g2l} operates on point clouds in a divide-and-conquer fashion and adopts a rotation residual estimator to estimate the residual between initial rotation and ground truth. 
CloudAAE~\cite{gao2021cloudaae} adopts an augmented autoencoder to improve the generalization of the network trained on synthetic depth data.

\textbf{Pose Estimation with RGB-D Data.}
When RGB images and depth images are employed individually for 6d pose estimation, both methods can achieve impressive performance. 
RGBD methods work with both RGB and depth information, tending to achieve a higher accuracy. 
PVNet~\cite{peng2019pvnet} can learn a vector field representation directed to the 2D keypoints.
DenseFusion~\cite{wang2019densefusion} utilises the 2D information within the embedding space to augment each 3D point and applies resulting colour depth space to predict 6D object pose. 
FFB6D~\cite{he2021ffb6d} presents a novel full flow bidirectional fusion network for representation learning from the RGBD image. 
KVNet~\cite{wang2023kvnet} estimates both the translation and rotation branch via Hough voting scheme. 
FoundationPose~\cite{wen2023foundationpose} designs a generative network to provide several pose hypotheses and selects the highest scoring pose by calculating the similarity.


\subsection{Graph Convolution Network (GCN)}
Due to the great representation power of graph structure,  GCN has achieved superior results in several tasks, especially human pose estimation~\cite{xu2021graph,wang2022uformpose} and remote sensing imagery~\cite{liu2023uncertainty}. 
Hence, some researchers draw the ideas from above tasks and try to introduce GCN into 3D vision domain.
PR-GCN~\cite{zhou2021pr} proposes a Multi-Modal fusion network base on GCN, which is applied to fuse the appearance and geometry features. GNN6D~\cite{yin2021graph} utilizes GCN to extract point cloud features and then attaches appearance feature to each node in graph.  DGECN~\cite{cao2022dgecn}  leverages geometry information to form Multi-Fusion feature, then generates 2D-3D correspondences by means of Encoder-Decoder architecture. 
Though demonstrate promising performance, these methods suffer from a lack of global exchange of geometry information, which results in ineffective adaptation to complicated scenarios. 

\subsection{Vision Transformer}
Transformer~\cite{vaswani2017attention} is first introduced as an attention-based framework in the field of Natural Language Processing (NLP). Thanks to the strong associative representational power of the attention mechanism, researchers have gradually applied it in computer vision tasks. 
Vit\cite{dosovitskiy2020image} is the pioneering work of Transformer in the field of 2D vision, which splits images into 16×16 patches and treats each patch as a token, and then leverages Transformer Encoder to extract image recognition features. 
DETR\cite{carion2020end} proposes a novel end-to-end object detection architecture and directly predicts the final set of detections by combining a common CNN with a transformer architecture.
Swin Transformer\cite{liu2021swin,liu2022swin} presents a hierarchical Transformer whose representation is computed with shifted windows. This scheme limits self-attention computation to non-overlapping local windows to bring greater efficiency. 
Wang et al.~\cite{wang2022hybrid} incorporates the Transformer architecture in the hybrid encoder (HE) to enable the model to capture the global context.

\begin{figure*}[htbp]
    \includegraphics[width=\textwidth]{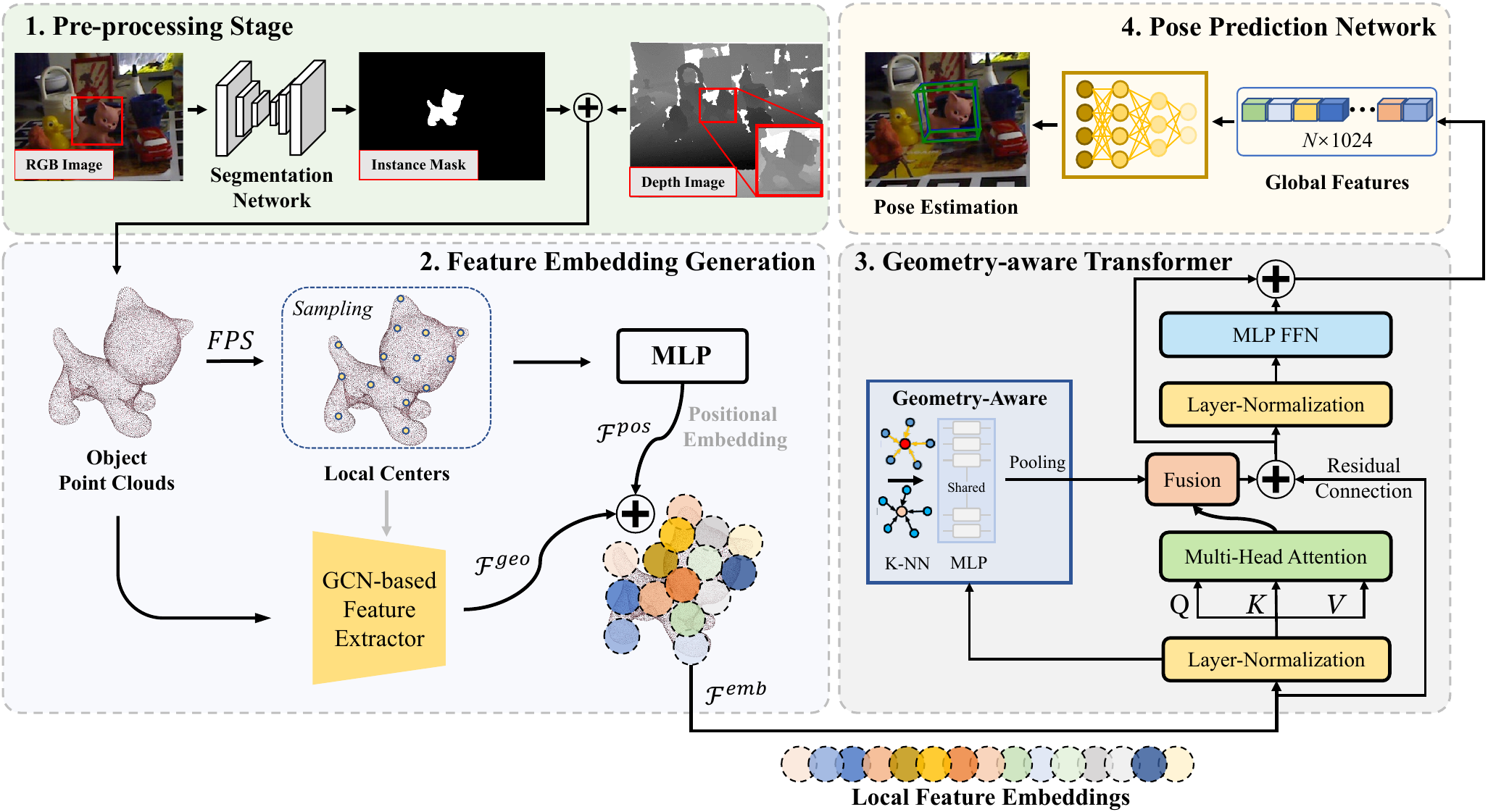}
    \caption{\textbf{Overview of TransPose.} The pre-processing stage obtaining the target object point cloud from the mask and depth image of the object with the camera internal reference transform(\emph{e.g., cat}). 
    The model employs a GCN-based feature extractor to obtain a local feature representation of the point cloud, and supplements it with a learnable positional encoding before passing into the Transformer Encoder. 
    Transformer block takes as input a local feature embeddings then fuses the results of the multi-head attention and geometry-aware module to produce a global feature representation
    The ultimate 6D pose estimation parameters are recovered by Pose Prediction Network.
    }\label{fig:TransPose}
\end{figure*}

For the 3D vision tasks, 
Zhou et al.~\cite{zhou20226} proposes local transformer and global transformer to better learn point cloud feature representations.
YOLOPose~\cite{amini2023yolopose} draws on the ideas of DETR, taking the learnable positional encoding to substitute the original fixed sine positional encoding.
Trans6D~\cite{zhang2023trans6d} designs the pure and hybrid transformer respectively and models the global dependencies among each patch via ViT-like Transformer Layers. 
Unfortunately, existing 3D vision Transformer methods consider less about the inductive bias modules that assume a constraining role in traditional visual models. And the lack of vision-related inductive bias probably reduces the accuracy and generalization ability of Transformer for processing vision tasks.
In contrast, our approach develops a geometry-aware module as inductive bias for the global Transformer Encoder, which form effective constraint for the proposed framework.

\section{Proposed Method}
\label{sec:Method}
Given an RGB-D image of object, the objective of 6D object pose estimation aims to determine the transformation between the target object coordinate system relative to the vision or laser sensor coordinate system.
Such transformation is represented by a matrix $T=[R|t]\in SE(3)$, which consists of translation $t\in \mathbb{R}^3$ and rotation $R\in SO(3)$ with three degrees of freedom respectively.
To better tackle this problem, the geometric and topological relations of the point cloud can provide assistance to the pose estimation algorithm in capturing discriminative feature.

\subsection{Overview}
\label{sec:Overview}
We propose TransPose, a novel 6D pose estimation framework with local and global geometry-aware feature extraction network, as shown in Figure~\ref{fig:TransPose}. As for the pre-processing stage, the instance mask of the target object is first obtained through a instance segmentation network. 
With the obtained mask, we can extract the object point cloud from depth images and take it as input of our proposed framework. The framework is mainly composed of three modules.
Specifically, \textbf{Feature Embeddings Generation} module utilizes designed graph convolution network to extract local point cloud feature, then flattens it and supplements it with a learnable positional encoding to form completed local feature embeddings. 
After that, we pass the feature into the \textbf{Geometry-aware Transformer Encoder}, which fuses the processing feature of multi-head attention mechanism and geometry-aware module to obtain global features. In this way, the output features will contain the geometric structure relationship in the high-dimensional feature.
Finally, the fusion point cloud output feature of the Transformer Encoder are fed into \textbf{Pose Prediction Network} to recover the final 6D pose estimation parameters.

\begin{figure}[htbp]
    \includegraphics[width=\columnwidth]{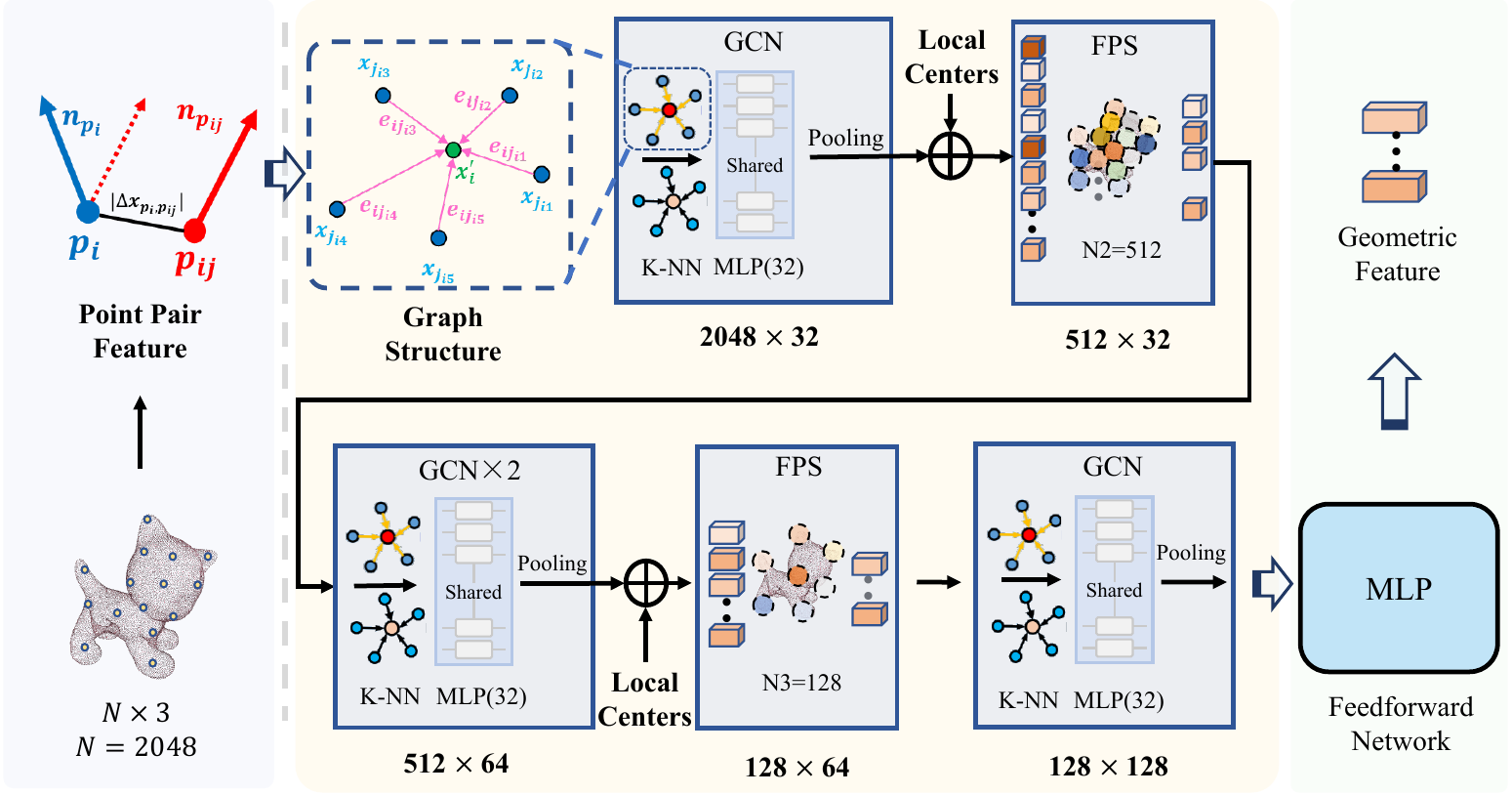}
    \caption{\textbf{The framework of GCN-based Local Feature Extractor}. The main network is composed of two modules: 
    \emph{(1)Graph Convolution}, which is the key component for extraction of local features. The module conducts K-Nearest Neighbor(K-NN) to determine the topology of the graph structure and converges neighborhood information to local centers via pooling.
    \emph{(2)Furthest Point Sampling (FPS)}, which exploits downsampling to reduce the number of point cloud sub-regions.
    The two modules connected at the string level are able to extract robust local features while boosting the efficiency of the algorithm.
    }\label{fig:GCNbase}
\end{figure}

\begin{figure}[htbp]
    \includegraphics[width=\columnwidth]{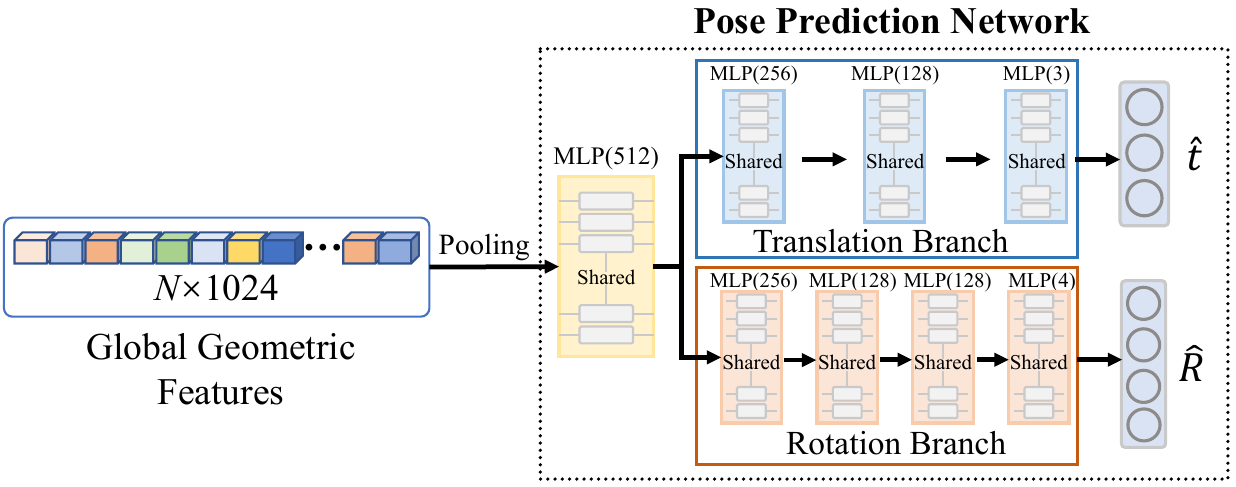}
    \caption{\textbf{Pose Prediction Network}. The network predicts the translation and rotation components through two decoupled branches respectively. Both sub-networks consist of a cascade of 1×1 convolution modules.
    }\label{fig:pre_network}
\end{figure}

\subsection{Point Cloud Feature Embeddings Generation}
\label{sec:Embeddings}

To take 3D point clouds suitable for Transformer Encoder, an trivial idea is to utilize the single point as a point cloud token and directly feeding the 3D coordinates of each point to the Transformer. 
However, since the computational complexity of the Transformer Encoder is quadratic to the sequence length, large-scale point cloud will lead to an unacceptable cost. 
On the other hand, unlike words in a sentence that include rich semantic information, the individual point of point cloud contain very limited information and cannot be directly applied the global self-attention mechanism.
Hence, it is crucial to divide the point cloud into different regions and extract local feature of point cloud.

To overcome the above limitations, we design a Feature Embedding Generation module to extract local geometry feature, and then flatten it and supplement it with a positional encoding to acquire a 3D point cloud encoding suitable for Transformer Encoder.
With the obtained the object point cloud, we first perform furthest point sampling(FPS) to sample fixed number N $\left\{ p_{1},p_{2}, \cdots ,p_{N}\right\}$as the center of the sub-regions. Then we need to extract the local feature around each point center. 
In recent years, several works of 3D human pose estimation based on graph convolution network have emerged~\cite{xu2021graph,wu2022local,wang2022uformpose}. These methods treat the joints of the human body as graph nodes to construct the human graph structure, and employ GCN  to obtain pose information for human pose estimation.
We notice that a point cloud can be also viewed as a special graph structure just like human skeleton, which comprises plenty of individual joint points. Therefore, it is possible to extract geometry information from point cloud data via graph convolution operations.
Inspired by that, we design a novel Feature Extractor base on Graph Convolution Network for 3D point cloud local feature extraction, which consists of the graph convolution block and Furthest Point Sampling block connected in series, as shown in Figure \ref{fig:GCNbase}.

Specifically, the graph convolution network block takes original point cloud and local center points as input, for center point $p_{i}$ of each local point cloud region, the initial local feature can be obtained in two steps: 
First, we perform K-Nearest Neighbor(K-NN) algorithm to determine the local domain area $\Psi(p_{i})$. 
Second, we construct the Point Pair Features (PPF)~\cite{drost2010model} between the local center point $p_{i}$ and the point $p_{ij}$ in domain area $\Psi(p_{i})$ to form edge of the graph structure:
\begin{equation}\label{equ:ppf}
\begin{split}
    PPF(p_{i},p_{ij})=( \angle (n_{p_{i}},\triangle x_{p_{i},p_{ij}}),  \angle (n_{p_{ij}}, \\
    \triangle x_{p_{i},p_{ij}}),\angle (n_{p_{i}},n_{p_{ij}}),|| \triangle x_{p_{i},p_{ij}}||_{2} )
\end{split}
\end{equation}
where $n_{p_{i}}$ and $n_{p_{ij}}$ represent the normal vectors of $p_{i}$ and $p_{ij}$.  $\triangle x_{p_{i},p_{ij}}=p_{ij}-p_{i}$ denotes the vectors between $p_{i}$ and $p_{ij}$. 
The Point Pair Features is a feature description with normal vector angle and Euclidean distance as a criterion, which has been shown to possess powerful geometry information representation ability~\cite{hodan2018bop}.
Moreover, the initial local feature of each point cloud region are mapped in high dimensions via the weight-shared MLP. Eventually, we perform pooling operation to aggregate the features to local centers to obtain outputs of the first layer.

In addition, the FPS block reduces the number of point cloud sub-regions via downsampling so as to improve the efficiency of the algorithm. 
These two blocks are connected in series to process the features.
After that, for center point $p_{i}$ of each point cloud local region, we can obtain the feature vector containing the local geometry information of the point cloud, denoted as $F_{i}^{geo}\in \mathbb{R}^{d_{in}}$.

Meanwhile, we map the original 3D coordinates to the same feature dimension $d_{in}$  as $F_{i}^{geo}$ via MLP to form a learnable position embedding for the center point $P_{i}^{pos}=\Phi _{a}\left ( p_{i} \right )$, where $a$ is the parameter of the MLP. The two vectors are added to obtain the final point cloud local feature embedding:
\begin{equation}\label{equ:femb}
    F_{i}^{emb}=P_{i}^{pos}+F_{i}^{geo}
\end{equation}
Suppose the final number of sampling points is $N$, then point cloud features embedding $\boldsymbol{F}^{emb}\in \mathbb{R}^{d_{in}}$ as the input feature for subsequent Transformer encoders.

\subsection{Geometry-aware Transformer}
\label{sec:Transformer}
Inspired by the structure for 2D images in \cite{dosovitskiy2020image}. We initially intended to feed the point cloud local features embeddings in layer normalization and multi-head attention module to obtain feature vectors in a single encoder. However, the multi-head attention mechanism of Transformer lacks inductive bias in traditional vision models, which is one of the key challenges for Transformer to be applied to the visual field.
Inductive bias is a critical concept in machine learning, which plays a critical role in determining the accuracy and generalization performance of a machine learning algorithm. 
Specifically,  it's the underlying knowledge or assumptions built into the learning algorithm that help it generalize from a limited set of training examples to new and unseen examples.
For instance, the Decision Tree algorithm has an inductive bias that the relationship between input and output data can be represented by a hierarchical structure.
CNNs views the information owns spatial locality that the parameter space can be reduced by sharing weights with sliding convolutions while RNNs considers time sequence information to stress the importance of order. 
The graph network believes that the similarity between the central node and the neighbor nodes will guide the flow of information better. Transformer is first applied in NLP, the original model does not have the inductive bias module naturally applicable to visual tasks. 

To enable the Transformer to better exploit the inductive bias about 3D geometry structure of point clouds, 
we apply the graph convolution network block mentioned in Section~\ref{sec:Embeddings} as the geometry-aware module to model geometry structural relationships in high-dimensional features, as shown in the Geometry-Aware Transformer block portion of Figure~\ref{fig:TransPose}.
\begin{table*}[htbp]
    \small
    \centering
    \setlength\tabcolsep{3pt}
    \caption{Ablation studies on the \textbf{LineMod Dataset} based on the \textbf{ADD(-S)} metric.  We use bold to represent the best results. Objects with bold name are symmetric.}
    \begin{tabular}{cc|ccccccccccccc|c}
    \toprule
     GCN &Geo-aware& ape & benchv & cam & can & cat & driller & duck & \textbf{eggbox} & \textbf{glue} & holep & iron & lamp & phone & MEAN\\
    \midrule
    -   & -         & 93.61         & 94.28          &97.15        & 96.25          & 99.2        & 97.12        & 92.77          & \textbf{100} & 99.61        & 96.66        & 96.22          & 96.55          & 97.79        & 96.71          \\
    \hline
    \checkmark   & -         & 95.99         & 97.08          & 98.04        & 96.47          & 99.00        & 98.22        & 96.24          & \textbf{100} & 99.61        & 97.91        & 97.94          & 98.08          & \textbf{100} & 98.06          \\
     \hline
    -   & \checkmark         & 93.99         & 96.12          & 98.24        & 96.46          & 99.5         & 97.62        & 95.12          & \textbf{100} & 99.52        & 96.86        & 98.26          & \textbf{99.23 }         & 97.89        & 97.60          \\
    \hline
    \checkmark   & \checkmark         & \textbf{98.1} & \textbf{99.03} & \textbf{100} & \textbf{99.01} & \textbf{100} & \textbf{100} & \textbf{99.06} & \textbf{100} & \textbf{100} & \textbf{100} & \textbf{98.97} &  99.04  & 99.04        & \textbf{99.40} \\
    \bottomrule
    \label{tab:LM_Ablation}
    \end{tabular}
\end{table*}
\begin{table}[htbp]
    \small
    \centering
    \setlength\tabcolsep{10pt}
    \caption{Ablation studies on the \textbf{Occlusion LineMod Dataset} based on the \textbf{ADD(-S)} metric.  We use bold to represent the best results. Objects with bold name are symmetric.}
    \begin{tabular}{r|c|c|c|c}
    \toprule
    GCN & - & \checkmark & - &\checkmark \\
    Geo-aware & - & - & \checkmark & \checkmark \\
    \midrule
    ape                      & 29.12                 & 44.85                 & 52.23                 & \textbf{57.01}        \\
    can                      & 20.73                 & 47.96                 & 35.8                  & \textbf{68.25}        \\
    cat                      & 9.21                  & 36.4                  & 35.98                 & \textbf{36.57}        \\
    driller                  & 48.26                 & 46.69                 & 61.36                 & \textbf{70.28}        \\
    duck                     & 27.09                 & 45.38                 & 45.65                 & \textbf{53.91}        \\
    \textbf{eggbox}          & 79.48                 & 77.59                 & 80.77                 & \textbf{80.92}        \\
    \textbf{glue}            & 73.86                 & 77.75                 & 75.08                 & \textbf{79.29}        \\
    holep                    & 57.61                 & 67.8                  & 68.63                 & \textbf{78.06}        \\
    \midrule[1pt]
    MEAN & 43.17                 & 55.55               & 56.94               & \textbf{65.54} \\
    \bottomrule
    \end{tabular}
    \label{tab:OccLM_Ablation}
\end{table}
The specific process is as follows:
\begin{equation}
\label{equ:sullp_to_perpend_normals_1}
\begin{split}
    F_{Attn}&=\mathbf{MHA}(\mathbf{LN}(F^{emb})),
    \\
    F_{GA}&=max(\mathbf{GA}(\mathbf{LN}(F^{emb}))),
    \\
    F&=Concat(F_{Attn} , F_{GA})+F^{emb},
    \\
    F_{out} &= \mathbf{FFN}(\mathbf{LN}(F)) + F
\end{split}
\end{equation}
where \emph{MHA}(·) is the Multi-Head Attention, \emph{LN}(·) indicates Layer-Normalization, \emph{GA}(·) denotes Geometry-Aware module and \emph{FFN}(·) is the feed-forward network.

Specifically, Transformer Encoder takes as input the local feature embeddings and feeds it into both the original multi-head self-attention process unit and geometry-aware module simultaneously after the layer normalization.
Different from the self-attention module that uses the feature similarity to capture the semantic relation, we propose to leverage the K-NN model of Geometry-Aware module to capture the geometric relation in the point cloud, and learn the local geometric structures by feature aggregation with a linear layer followed by the max pooing operation.


The geometric feature and semantic feature are then concatenated and mapped to the original dimensions to form the output.
Following this, a dimensional reduction mapping will be applied to restore the features to original dimensions. 
These reconstructed features will then be concatenated with the input features through residuals, culminating in the derivation of the ultimate global features of the point cloud.
The geometry-aware module is introduced as an inductive bias to achieve an effective combination of global semantic features and local geometric features, forming an appropriate constraint for point cloud learning.

\subsection{Pose Prediction Network}
\label{sec:Predict_Network}

With the obtained global geometric features, we can predict the final pose parameters of objects.
Similar with~\cite{wen2020se}, we first perform a pooling operation and then feed them into the designed pose prediction network structure based on translation and rotation decoupling to recover 6D pose estimation parameters, as shown in Figure~\ref{fig:pre_network}.

For the 3D translation, the original point cloud is first translated to the local canonical coordinate and then the subsequent feature extraction and pose predict. We define the origin of the local normalized coordinate system as the barycenter of the original point cloud.
\begin{equation}\label{equ:zuobiao}
    \bar{X}=\left ( \bar{x},\bar{y},\bar{z} \right )=\frac{1}{N}\left ( \sum_{i=1}^{N}x_{i},\sum_{i=1}^{N}y_{i},\sum_{i=1}^{N}z_{i} \right )
\end{equation}
The expected output of the translation prediction network is the difference between the true value $t$ and the point cloud center $\bar{X}$, that is $t-\bar{X}$.

For the regression of the rotation, we utilize quaternion as the representation of the network predicted rotation amount to avoid the Gimbal Lock problem in Euler rotation. 
The quaternion consists of a scalar and a vector. In this paper, the rotation quaternion is represented in the form $\mathbf{q}=q_{3}+q_{0}\mathbf{i}+q_{1}\mathbf{j}+q_{2}\mathbf{k}$. 
Meanwhile, the four-dimensional vectors of network predictions need to be standardized to ensure that $\left\|\mathbf{q}\right\|=q_{0}^{2}+q_{1}^{2}+q_{2}^{2}+q_{3}^{2}=1$. 
Hence, the corresponding rotation matrix $R(\mathbf{q})$ can be obtained:
\begin{equation}\label{equ:rotation}
    \left[\begin{array}{ccc}
1-2q_{1}^{2}-2q_{2}^{2} & 2q_{0}q_{1}-2q_{2}q_{3} & 2q_{0}q_{2}-2q_{1}q_{3} \\
2q_{0}q_{1}+2q_{2}q_{3} & 1-2q_{0}^{2}-2q_{2}^{2} & 2q_{1}q_{2}-2q_{0}q_{3} \\
2q_{0}q_{2}-2q_{1}q_{3} & 2q_{1}q_{2}+2q_{0}q_{3} & 1-2q_{0}^{2}-2q_{1}^{2} 
\end{array}\right]
\end{equation}

\section{Experiments}
\label{sec:Experiments}
\subsection{Datasets and Metrics}
\label{sec:dataset_metrics}
We evaluate our method on three mainstream benchmark datasets.

\textbf{LineMod}~\cite{hinterstoisser2011gradient} is a dataset consist of 13 low-texture living objects sequences, each containing about 1.2K groups of aligned RGB images and depth images with corresponding camera parameters. The texture-less objects, cluttered scenarios, and varying lighting make this dataset challenge. Following the domain consensus~\cite{brachmann2016uncertainty}, about 15\% of each category of object is selected for training, and the remaining 85\% of the data is used for testing.

\begin{table*}[htbp]
    \small
    \centering
    \setlength\tabcolsep{4pt}
    \renewcommand\arraystretch{1.2} 
    \caption{Comparison of performance with GCN-based and Transformer-based methods on the \textbf{LineMod Dataset}.  We use bold to represent the best results, and underline the second-best results. ($^*$)indicates the method performs refinement process. (${_g}$) represents GCN-based method. (${_T}$) denotes Transformer-based method. Objects with bold name are symmetric.}
    \begin{tabular}{r|ccccccccccccc|c}
    \toprule
     Methods & ape & benchv & cam & can & cat & driller & duck & \textbf{eggbox} & \textbf{glue} & holep & iron & lamp & phone & MEAN\\
    \midrule
    GNN6D$^{*}_{g}$\cite{yin2021graph}         & 82.47         & 97.63         & 88.43        & 95.17         & 93.41        & 94.44        & 86             & \uline{99.9}        & \uline{99.9}        & 86.77        & 91.52         & 97.69         & 94.81         & 92.95          \\
    \hline
   Trans6D+$^{*}_{T}$\cite{zhang2023trans6d}          & 88.3          & \uline{99.4} & 97.8         & \textbf{99.1} & 93.2         & \uline{99.5} & 87.8           & \textbf{100} & 99.8         & 96.7         & \textbf{99.9} & \textbf{99.7} & \textbf{99.5} & 96.9                                   \\
     \hline
    PR-GCN$_{g}$\cite{zhou2021pr}          & \uline{ 97.6}          & \uline{ 99.2}       &99.4       & 98.4          &98.7         & 98.8         & \uline{98.9}      & \uline{99.9}         & \textbf{100} & 99.4         & 98.5          & 99.2          & 98.4          & 98.9          \\
    \hline
    Zhou et al.$_{T}$\cite{zhou20226}          & 97.52         &  \textbf{99.41}       & \uline{99.41}       &  \uline{99.21}         &\uline{99.9}         & 99.31        & 98.12      & \textbf{100}         & \textbf{100} & \uline{99.52}        & 98.26          & 99.40         & 98.66      & \uline{99.13}        \\
    \hline
    \textbf{Ours}         & \textbf{98.1} & 99.03         & \textbf{100} &99.01         & \textbf{100} & \textbf{100} & \textbf{99.06} & \textbf{100} & \textbf{100} & \textbf{100} & \uline{98.97}         & 99.04         & \uline{99.04}        & \textbf{99.40} \\
    \bottomrule
    \end{tabular}
    \label{tab:LM_same_compare}
\end{table*}

\begin{table}[htbp]
    \small
    \centering
    \setlength\tabcolsep{1.5pt}
    \caption{Comparison of performance with GCN-based and Transformer-based methods on the \textbf{Occlusion LineMod Dataset}.  We use bold to represent the best results, and underline the second-best results. ($^\dagger$) indicates the method utilizes additional synthetic data for training. ($^*$)indicates the method performs refinement process. (${_g}$) represents GCN-based method. (${_T}$) denotes Transformer-based method. Objects with bold name are symmetric.}
    \begin{tabular}{r|c|c|c|c|c}
    \toprule
    Methods & \Methods{Trans6D+$^{*}_{T}$}{\cite{zhang2023trans6d}} & \Methods{DGECN$_{g}$}{\cite{cao2022dgecn}} &\Methods{Zhou et al.$^{\dagger}_T$}{\cite{zhou20226}}&  \Methods{PR-GCN$^{\dagger}_{g}$}{\cite{zhou2021pr}}  & \textbf{Ours} \\
    \midrule
    ape                      & 36.9          & \uline{50.3}  & 42.03  & 40.2                    & \textbf{57.01}           \\
    can                      & \textbf{91.6} & 75.9     & 67.48      & \uline{76.2}             & 68.25                    \\
    cat                      & \uline{42.5}          & 26.4      & 33.13     &\textbf{57}      & 36.57                    \\
    driller                  & \uline{70.8}          & 77.5     & 63.58      & \textbf{82.3}    & 70.28                    \\
    duck                     & 41.1          & \textbf{54.2} & 45.44& 30                      & \uline{53.91}              \\
    \textbf{eggbox}         & 56.3          & 57.8       & \uline{77.07}   &68.2                    & \textbf{80.92}  \\
    \textbf{glue}            & 62            & 66.9    & \uline{78.13}      & 67                      & \textbf{79.29}           \\
    holep                    & 61.9          & 60.2    & 74.29       &\textbf{97.2}          & \uline{78.06}              \\
    \midrule[1pt]
    MEAN & 57.9          & 58.7    & 60.14    &\uline{65}                & \textbf{65.54} \\
    \bottomrule
    \end{tabular}
    \label{tab:OccLM_same_compare}
\end{table}

\textbf{Occlusion LineMod}~\cite{brachmann2016uncertainty} dataset is an extension of the original LineMod dataset. The 8 objects are selected and annotated with 6DoF poses of a single object from LineMod dateset. Each image in this dataset consists of multi annotated objects, which are heavily occluded. In extreme cases, the observable part of the target surface is less than 10\% of the total foreground area.

\textbf{YCB-Video Dataset}~\cite{xiang2017posecnn} features 21 objects of varying shape and texture different from YCB object set~\cite{calli2015ycb}.
The dataset contains 92 real RGB-D video sequences in total, where each video shows a subset of the 21 objects in different indoor scenarios. Following the prior work~\cite{xiang2017posecnn}, we select 80 sequences for training and 2,924 key frames from the remaining 12 sequences for testing.
Furthermore, the training set includes 80K synthetic images, allowing the YCB-V dataset to encompass a wider range of challenging scenarios, such as changing lighting conditions, occlusions, and the presence of image noise.

We adopt the the commonly used metrics Average 3D Distance(ADD-S) Metric~\cite{hinterstoisser2012model} and ADD-S for evaluation.  
The ADD metric meatures the average deviation between objects transformed by the predicted and the ground truth pose, and judge whether the accuracy of distance is less than a certain fraction of the object's diameter($e.g.$ ADD-0.1d), defined as follow:
\begin{equation}\label{equ:eADD}
    e_{ADD}=\underset{x\in M}{avg}\left\|(Rx_{i}+t)-(\hat{R}x_{i}+\hat{t}) \right\|_{2}
\end{equation}
where $x$ denotes a vertex in object $M$, $R$, $t$ denote the ground truth and $\hat{R}$, $\hat{t}$ denote the predicted pose. For symmetric objects, ADD-S computes the average distance to the closest model point:
\begin{equation}\label{equ:eADD-S}
    e_{ADD-S}=\underset{x_{2}\in M}{avg}\underset{x_{1}\in M}{min}\left\|(Rx_{1}+t)-(\hat{R}x_{2}+\hat{t}) \right\|_{2}
\end{equation}

\subsection{Ablation Studies}
\label{sec:Ablation_studies}
In this section, we investigate the effectiveness of graph convolution network block in local feature extractor as well as the geometry-aware module in Transformer Encoder.
For graph convolution network block, we keep the point cloud downsampling process to guarantee the equality of comparison and prevent the large computational complexity. 
Subsequently, we utilizes a simple convolution layer(Linear, BatchNorm and ReLU) to replace the graph convolution network block, and the number of simple convolution layer's output channels is corresponding to the point cloud sampling result.
As for geometry-aware module,  the ablation process without the geometry-aware module means using the regular transformer encoder to process the Local Feature Embeddings.
Then we perform ablation experiments on the LineMod and Occlusion LineMod dataset. We use bold to indicate the best results.

The quantitative results of ablation studies on LineMod dataset are shown in Table~\ref{tab:LM_Ablation}. The last column in table displays the average accuracy results, from which we can clearly see that the baseline model achieves promising accuracy of 96.71\%, which is already a competitive result compared to other methods.
Despite that, our graph convolution network block and geometry-aware module boost the baseline by 1.35\% and 0.89\%, respectively. 
In particular, the complete model with both  graph convolution network block and geometry-aware module achieves a higher accuracy on almost every single object.  The overall average accuracy is 99.4\% , which indicates that the proposed module contributes to object pose estimation.
In addition, the results of ablation studies on Occlusion LineMod dataset are shown in Table~\ref{tab:OccLM_Ablation}, which exhibits a more obvious parallel conclusion.  
The baseline model can only achieve 43.17\% on the more challenging Occlusion LineMod dataset.  Our graph convolution network block and geometry-aware module bring a 12.38\% and 13.77\% performance improvement, respectively.  The complete model achieves higher accuracy of 65.54\%,  which further demonstrates the effectiveness of the proposed module.

\begin{table*}[htbp]
    \small
    \centering
    \setlength\tabcolsep{3pt}
    \caption{Quantitative comparison on the \textbf{LineMod Dataset} based on the \textbf{ADD(-S)} metric. ($^\dagger$) indicates the method utilizes additional synthetic data for training. We use bold to represent the best results for each modality. And the overall best results are underlined. Objects with bold name are symmetric.}
    \begin{tabular}{r|cccc|cccc|cccc}
    \toprule
    Input & \multicolumn{4}{c|}{RGB} & \multicolumn{4}{c|}{D} & \multicolumn{4}{c}{RGBD} \\ 
    \midrule
    Methods & \Methods{Pix2pose}{\cite{park2019pix2pose}} & \Methods{PVNet$^{\dagger}$}{\cite{peng2019pvnet}} & \Methods{CDPN$^{\dagger}$}{\cite{li2019cdpn}} & \Methods{DPOD$^{\dagger}$}{\cite{zakharov2019dpod}} & \Methods{Cloudpose}{\cite{gao20206d}} & \Methods{CloudAAE$^{\dagger}$}{\cite{gao2021cloudaae}} & \Methods{G2L-Net}{\cite{chen2020g2l}} & \textbf{Ours} & \Methods{KVNet$^{\dagger}$}{\cite{wang2023kvnet}}& \Methods{Uni6D$^{\dagger}$}{\cite{jiang2022uni6d}}   & \Methods{PVN3D$^{\dagger}$}{\cite{he2020pvn3d}} & \Methods{FFB6D$^{\dagger}$}{\cite{he2021ffb6d}} \\ 
    \midrule
    ape & 58.1 & 43.62 & 64.4 & \textbf{87.73} & 58.3 & 92.5 & 96.8    & \textbf{98.10}  & 93.2 & 93.71   & 97.3 & \uline{\textbf{98.4}} \\
    benchv & 91.0 & \textbf{99.90} & 97.8 & 98.45 & 65.6 & 90.8 & 96.1   & \textbf{99.03} & 97.1  & 99.81  & 99.7 & \uline{\textbf{100}} \\
    cam & 60.9 & 86.86 & 91.7 & \textbf{96.07} & 43.0 & 85.7 & 98.2   & \uline{\textbf{100}}  & 96.4 & 95.98  & 99.6 & \textbf{99.9} \\
    can & 84.4 & 95.47 & 95.9 & \textbf{99.71} & 84.7 & 95.1 & 98.0   & \textbf{99.01} & 97.7 & 99.01  & 99.5 & \uline{\textbf{99.8}} \\
    cat & 65.0 & 79.34 & 83.8 & \textbf{94.71} & 84.6 & 96.8 & 99.2   & \uline{\textbf{100}}  & 98.4 & 98.10  &  99.8 & \textbf{99.9} \\
    driller & 76.3 & 96.43 & 96.2 & \textbf{98.8} & 83.3 & 98.7 & 99.8   & \uline{\textbf{100}}  & 93.8 & 99.11  &  99.3 & \uline{\textbf{100}} \\
    duck & 43.8 & 52.58 & 66.8 & \textbf{86.29} & 43.2 & 84.4 & 97.7   & \uline{\textbf{99.06}}  & 95.5 & 89.95  &  98.2 & 98.4 \\
    \textbf{eggbox} & 96.8 & 99.15 & 99.7 & \textbf{99.91} & 99.5 & 99.2   & \uline{\textbf{100}}   & \uline{\textbf{100}} & \uline{\textbf{100}}& \uline{\textbf{100}}  & 99.8 & \uline{\textbf{100}} \\
    \textbf{glue} & 79.4 & 95.66 & \textbf{99.6} & 96.82 & 98.8 & 98.7 & \uline{\textbf{100}}   & \uline{\textbf{100}} & 99.9  & 99.23   & \uline{\textbf{100}} & \uline{\textbf{100}} \\
    holep & 74.8 & 81.29 & 85.8 & \textbf{86.87} & 72.1 & 85.3 & 99.0   & \uline{\textbf{100}}  & 93.2 & 90.20   & \textbf{99.9} & 99.8 \\
    iron & 83.4 & 98.88 & 97.9 & \uline{\textbf{100}} & 70.3 & 91.4 & \textbf{99.3}   & 98.97  & 98.6 & 99.49   & 99.7 & \textbf{99.9} \\
    lamp & 82.0 & \textbf{99.33} & 97.9 & 96.84 & 93.2 & 86.5 & \textbf{99.5}   & 99.04  & 98.9 & 99.42  & 99.8 & \uline{\textbf{99.9}} \\
    phone & 45.0 & 92.41 & 90.8 & \textbf{94.69} & 81.0 & 97.4 & 98.9   & \textbf{99.04} & 97.3  & 97.41   & 99.5 & \uline{\textbf{99.9}} \\ 
    \midrule
    MEAN & 72.4 & 86.27 & 89.9 & \textbf{95.15} & 75.2 & 92.5 & 98.7   & \textbf{99.40} & 96.9  & 97.03  & 99.4 & \uline{\textbf{99.7}} \\
    \bottomrule
    \end{tabular}
    \label{tab:LM_data}
\end{table*}

\begin{table*}[htbp]
    \small
    \centering
    \setlength\tabcolsep{3.5pt}
    \caption{Quantitative comparison on the \textbf{Occlusion LineMod Dataset} based on the\textbf{ ADD(-S)} metric. We use bold to represent the best results for each modality.  And the overall best results are underlined. ($^\dagger$) indicates the method utilizes additional synthetic data for training. Objects with bold name are symmetric.}
    \begin{tabular}{r|cccc|ccp{0.8cm}|cccc}
    \toprule
    Input & \multicolumn{4}{c|}{RGB}  & \multicolumn{3}{c|}{D}      & \multicolumn{4}{c}{RGBD}       \\ 
    \midrule
    Methods  & \Methods{Pix2pose}{\cite{park2019pix2pose}} & \Methods{ER-Pose}{\cite{yang2023er}}  & \Methods{GDR-Net$^{\dagger}$}{\cite{wang2021gdr}} &  \Methods{ZebraP$^{\dagger}$}{\cite{su2022zebrapose}}     & \Methods{CloudAAE$^{\dagger}$}{\cite{gao2021cloudaae}} &\Methods{Zhou et al.$^{\dagger}$}{\cite{di2021so}} & \textbf{Ours}   & \Methods{Uni6D$^{\dagger}$}{\cite{jiang2022uni6d}} & \Methods{PVN3D$^{\dagger}$}{\cite{he2020pvn3d}} & \Methods{GNN6D$^{*}_{g}$}{\cite{yin2021graph}} & \Methods{FFB6D$^{\dagger}$}{\cite{he2021ffb6d}} \\ 
    \midrule
    ape & 22.0 & 25.9  & 46.8  &\uline{\textbf{57.9}}  & - & 42.03& \textbf{57.01} & 32.99 & 33.9 & \textbf{48.53} & 47.2 \\
    can & 44.7 & 72.1  & 90.8  &\uline{\textbf{95.0}}  & -& 67.48 & \textbf{68.25} & 51.04 & \textbf{88.6} & 82.76 & 85.2 \\
    cat & 22.7 & 25.3  & 40.5  &\textbf{60.6}  & -& 33.13 & \textbf{36.57} & 4.56 & 39.1 & \uline{\textbf{62.79}} & 45.7 \\
    driller & 44.7 & 72.9 & 82.6  &\uline{\textbf{94.8}}  & -& 63.58 & \textbf{70.28} & 58.4 & 78.4 & \textbf{84.94} & 81.4 \\
    duck & 15.0 &35.8   & 46.9  & 64.5 & -& 45.44 & \textbf{53.91} & 34.8 & 41.9 & 43.98 & \textbf{53.9} \\
    \textbf{eggbox} & 25.2 & 48.7 & 54.2  & \textbf{70.9} & -& 77.07 &\textbf{\uline{80.92}}  & 1.73 & \textbf{80.9} & 61.31 & 70.2 \\
    \textbf{glue} & 32.4 & 58.8 & 75.8 & \uline{\textbf{88.7}} & -  & 78.13&\textbf{79.29} & 30.16 & \textbf{68.1} & 65.74 & 60.1 \\
    holep & 49.5 & 47.4 & 60.1 & \textbf{83.0} & - & 74.29 &\textbf{78.06} & 32.07 & 74.7 &73.02  &\uline{\textbf{85.9}} \\
    \midrule
    MEAN & 32.0 & 48.3 & 62.2 &\uline{\textbf{76.9}}  & 58.9  & 60.14& \textbf{65.54} & 30.71 & 63.2 & 65.38 & \textbf{66.2} \\ 
    \bottomrule
    \end{tabular}
    \label{tab:Occ_data}
    \vspace{-0.2cm} 
\end{table*}

\subsection{Comparison with the Same Type Methods}
\label{sec:same_type_comparison}
We also compare our method with other GCN-based methods and Transformer-based methods on LineMod and Occlusion LineMod dataset. The results are respectively exhibited in Table~\ref{tab:LM_same_compare} and Table~\ref{tab:OccLM_same_compare}. 
Our method displays competitive performance on the majority of objects compared to other methods, and overall accuracy outperforms other methods in both datasets.
Compared with GCN-based methods, we achieve a higher accuracy than GNN6D~\cite{yin2021graph} and PR-GCN~\cite{zhou2021pr} by 6.47\% and 0.5\% on LineMod dataset. For Occlusion LineMod dataset, we outperform above two methods and DGECN~\cite{cao2022dgecn} by 0.16\%, 0.54\% and 6.84\%. 
Compared with Transformer-based methods, we exceed Trans6D~\cite{zhang2023trans6d} and Zhou et al.~\cite{zhou20226} in both LineMod and Occlusion LineMod datasets.
Since the rest works of this series~\cite{amini2022t6d,beedu2022video,amini2023yolopose} only evaluate the YCB-Video dataset, we will include the comparison in Table~\ref{tab:YCB_data}.

Notably, our method neither performs refinement nor utilizes additional synthetic data for training, which still outperforms other methods that utilize these optimization means.
Experimental results indicate that our method outperforms other similar GCN-based and Transformer-based methods, indicating our pose estimation pipeline is effective.


\begin{table*}[htbp]
    \small
    \centering
    \setlength\tabcolsep{2pt}
    \caption{Quantitative comparison on the \textbf{YCB-Video Dataset} based on the \textbf{ADD-S AUC} metric.  We use bold to represent the best results for each modality. And the overall best results are underlined. (${_g}$) represents GCN-based method. (${_T}$) denotes Transformer-based method. Objects with bold name are symmetric.}
    \begin{tabular}{r|cccc|ccp{0.8cm}|cccc}
    \toprule
    Input & \multicolumn{4}{c|}{RGB} & \multicolumn{3}{c|}{D} & \multicolumn{4}{c}{RGBD} \\ 
    \midrule
    Methods & \Methods{PoseCNN}{\cite{park2019pix2pose}} & \Methods{VideoP${_T}$}{\cite{beedu2022video}} &  \Methods{ZebraP}{\cite{su2022zebrapose}} & \Methods{GDR-Net}{\cite{wang2021gdr}} & \Methods{G2L-Net}{\cite{chen2020g2l}} &\Methods{Zhou et al${_T}$}{\cite{zhou20226}} & \textbf{Ours} & \Methods{DGECN${_g}$}{\cite{cao2022dgecn}} & \Methods{DenseF}{\cite{wang2019densefusion}}  &  \Methods{FFB6D}{\cite{he2021ffb6d}} &\Methods{FoundP}{\cite{wen2023foundationpose}} \\ 
    \midrule
    \textbf{002\_master\_chef\_can} & 83.9 & 93.3  & 93.7 & \textbf{96.3} & 94.0& 95.1 & \textbf{95.67} &- & 95.3  & 96.3 & \uline{\textbf{96.9}}\\
    \textbf{003\_cracker\_box} & 76.9 & 78.2& 93.0 & \textbf{97.0} & 88.7 & 91.1& \textbf{92.13} & - & 92.5 & 96.3 &\uline{\textbf{97.5}}\\
    004\_sugar\_box & 84.2 & 82.5& 95.1 & \uline{\textbf{98.9}} & 96.0 & 96.03& \textbf{96.91} & - & 95.1  & \textbf{97.6} &97.5\\
    \textbf{005\_tomato\_soup\_can} & 81.0 & 91.1& 94.4 & \textbf{96.5} & 86.4& \textbf{95.39} & 93.87 & - & 93.8 &  95.6 &\uline{\textbf{97.6}}\\
    006\_mustard\_bottle & 90.4 & 91.8& 96.0 & \uline{\textbf{100}} & 95.9 & \textbf{97.01}& 96.98 & - & 95.8   & 97.8 &\textbf{98.4}\\
   \textbf{ 007\_tuna\_fish\_can} & 88.0 & 94.0& 96.9 & \uline{\textbf{99.4}} & 96.0 & \textbf{97.11}& \textbf{97.11} & - & 95.7 & 96.8 &\textbf{97.7}\\
    008\_pudding\_box & 79.1 & 90.3& \textbf{97.2} & 64.6 & 93.5 & 91.05& \textbf{95.45} & - & 94.3 &  97.1 &\uline{\textbf{98.5}}\\
    009\_gelatin\_box & 87.2 & 93.1 & 96.8 & \textbf{97.1} & 96.8&95.88 & \textbf{97.24} & - & 97.2  & 98.1 &\uline{\textbf{98.5}}\\
   \textbf{ 010\_potted\_meat\_can} & 78.5 & 89.3   & \textbf{91.7} & 86.0 & 86.2&\textbf{91.42} & 90.06 &- & 89.3  & 94.7&\uline{\textbf{96.6}} \\
    011\_banana & 86.0 & 81.3  & 92.6 & \textbf{96.3} & 96.3&95.91& \textbf{97.24} & - & 90.0  &  \textbf{97.2} &\uline{\textbf{98.1}}\\
    019\_pitcher\_base & 77.0 & 90.6 & 96.4 & \uline{\textbf{99.9}} & 91.8&\textbf{97.02} & 96.63 & - & 93.6  & 97.6 &\textbf{97.9}\\
    021\_bleach\_cleanser & 71.6 & 88.4 & 89.5 & \textbf{94.2} & 92.0 &93.45& \textbf{94.12} & - & 94.4   & 96.8 &\uline{\textbf{97.4}}\\
   \textbf{ 024\_bowl} & 69.6 & 78.8  & 37.1 & \textbf{85.7} & 86.7&94.47 & \textbf{95.35} & - & 86.0 & \uline{\textbf{96.3}} & 94.9\\
    025\_mug & 78.2 & 91.7 & 96.1 & \uline{\textbf{99.6}} & 95.4&96.71 & \textbf{96.78} & - & 95.3 & \textbf{97.3} &96.2\\
    035\_power\_drill & 72.7& 82.7  & 95.0 & \textbf{97.5} & \textbf{95.2}&93.42 & 94.80 & - & 92.1 & \textbf{97.2} &\uline{\textbf{98.0}}\\
   \textbf{ 036\_wood\_block} & 64.3 & 68.6 & \textbf{84.5} & 82.5 & 86.2 &87.5& \textbf{89.58} &- & 89.5 &  92.6 &\uline{\textbf{97.4}}\\
    037\_scissors & 56.9 & \textbf{92.5} & 63.8 & 60.8& 83.8  &89.11& \textbf{90.93} & - & 90.1  & 97.7 &\uline{\textbf{97.8}}\\
    \textbf{040\_large\_marker} & 71.7 & 84.2  & 80.4 & \textbf{88.0} & \textbf{96.8}&94.51& 95.84 & - & 95.1  & 96.6 &\uline{\textbf{98.6}}\\
    \textbf{051\_large\_clamp} & 50.2 & 81.8  & 85.6 & \textbf{89.3} & \textbf{94.4}&73.59 & 74.57 & - & 71.5 &  96.8 &\uline{\textbf{96.9}}\\
    \textbf{052\_extra\_large\_clamp} & 44.1 &60.6 & 92.5 & \textbf{93.5} & \textbf{92.3}&83.19 & 69.49 & - & 70.2 &  96.0 &\uline{\textbf{97.6}}\\
    \textbf{061\_foam\_brick} & 88.0 &92.7 & 95.3 & \textbf{96.9} & 94.7&94.4 & \textbf{96.09} & - & 92.2 &  97.3&\uline{\textbf{98.1}}\\
    \midrule
    MEAN & 75.8 & 85.3 &90.1  & \textbf{91.6} & 92.4&92.51 & \textbf{92.71} & 90.9 & 91.2 & 96.6 &\uline{\textbf{97.4}}\\
    \bottomrule
    \end{tabular}
    \label{tab:YCB_data}
    \vspace{-0.2cm} 
\end{table*}

\subsection{Evaluations on Benchmark Datasets}
\label{sec:evaluation}
\emph{\textbf{1) Evaluations on LineMod Dataset}}: The quantitative results of our \textbf{TransPose} and state-of-the-art method on LineMod are exhibited in Table~\ref{tab:LM_data}. According to the different modalities used in the pose inference phase, we divide these methods into three categories to make the comparison clearer. The best results for each modality are in bold and the overall best results are underlined.

As the table shows, the average accuracy of our method is 99.4\%, exceeding all the RGB-based and the depth-based methods, as well as most RGBD-based methods. Concretely, the accuracy of our method is 4.25\% higher than the best RGB-based method DPOD~\cite{zakharov2019dpod}. 
Among methods based on depth image, our method is 0.4\% ahead of the second place G2L-Net~\cite{chen2020g2l}. It is noteworthy that both approaches use only real data for training instead of additional synthetic data. Our model still achieves a slightly better performance with saturated recall. 
Besides, our method is 6.9\% higher than CloudAAE~\cite{gao2021cloudaae}, which utilizes additional synthetic data to obtain better performance.
Compared with RGBD based methods, our model outperforms most methods and achieves the same accuracy as PVN3D~\cite{he2020pvn3d}, only slightly behind FFB6D~\cite{he2021ffb6d}. It is worth noting that PVN3D~\cite{he2020pvn3d} and FFB6D~\cite{he2021ffb6d} use large-scale synthetic dataset(20K per object) to achieve the current accuracy.
To summarize, our method can achieves competitive results without using RGB image in pose inference stage compared with other methods, exhibiting the effectiveness of our pipeline.

\emph{\textbf{2) Evaluations on Occlusion LineMod Dataset}}: To verify the robustness of our model for inter-object occlusion situations, we report the quantitative results on the Occlusion LineMod dataset.
We follow prior works~\cite{peng2019pvnet,park2019pix2pose} and directly utilize the pre-trained model on the LineMod dataset for testing. The quantitative results are displayed in Table~\ref{tab:Occ_data}. Analogously, we divide these methods based on the modality.

As the Table~\ref{tab:Occ_data} shows, our method achieves a fairly competitive average accuracy of 65.54\%, which is higher than most methods. Specifically, our method is higher than depth-based CloudAAE~\cite{gao2021cloudaae} and Zhou et al,~\cite{zhou20226} by 6.64\% and 5.4\%, our method also outperforms RGB-based methods Pix2pose~\cite{park2019pix2pose}, ER-Pose~\cite{yang2023er}, GDR-Net~\cite{wang2021gdr} by margins of 33.54\%, 17.24\% and 3.34\% , respectively. 
Compared with RGBD-based methods, we surpass Uni6D~\cite{jiang2022uni6d}, PVN3D~\cite{he2020pvn3d} by 34.81\% and 2.34\%, and is on par with FFB6D~\cite{he2021ffb6d}. Most of above methods utilize additional synthetic data for training and has better viewpoint coverage.
Nevertheless, the accuracy of our method is behind state-of-the-art ZebraPose~\cite{su2022zebrapose}. The main reason is that method employs a dense prediction approach, which will be more advantageous when dealing with occlusions. 
Additionally, ZebraPose~\cite{su2022zebrapose} utilizes PBR~\cite{hodavn2019photorealistic} dataset, which has a total of 400K images. The large-scale images dataset with a high degree of visual realism allows to reduce the domain gap between different dataset.
In summary, our method still exhibits promising performance in the more challenging Occlusion LineMod dataset. 

\begin{figure*}[htbp]
    \includegraphics[width=\textwidth]{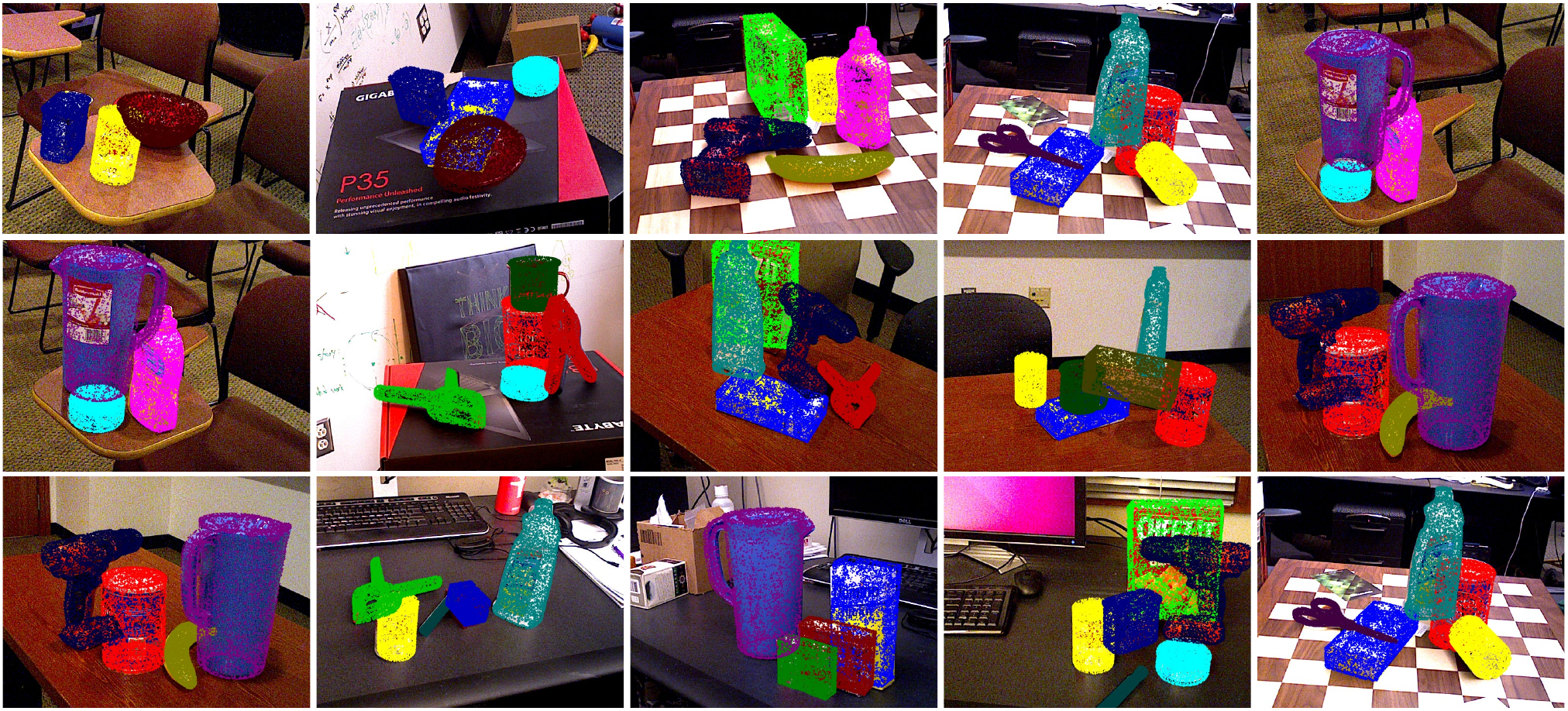}
    \caption{Qualitative results on the YCB-Video dataset. We project the predicted poses as point cloud onto each model in the RGB image. Different objects are depicted by different colors. 
        }\label{fig:YCB_visual}
\end{figure*}

\emph{\textbf{3) Evaluations on YCB-Video Dataset}}: Table~\ref{tab:YCB_data} displays the quantitative results on YCB-Video dataset. In practice, most of the objects in dataset are actually geometrically symmetric but asymmetrical in appearance. Since we only apply depth data to estimate 6D pose, our method can not distinguish the appearance differences. Hence, we report the results based on the ADD-S AUC metric following PoseCNN~\cite{xiang2017posecnn}.

As shown in the table, the average accuracy of our method is 92.7\%, surpassing most of listed methods. Specifically, our model is able to outperform RGB-based methods PoseCNN~\cite{park2019pix2pose}, ZebraPose~\cite{su2022zebrapose} and GDR-Net~\cite{wang2021gdr} by margins of 16.91\%, 2.61\% and 1.11\% respectively. In addition, we outperform Transformer-based method VideoPose~\cite{beedu2022video} by 7.41\%.

For depth-based method, we are higher than G2L-Net~\cite{chen2020g2l} and Zhou et al,~\cite{zhou20226} by 0.31\% and 0.2\%. In particular, we achieve a higher accuracy than RGB-D methods DGECN~\cite{cao2022dgecn} and DenseFusion~\cite{wang2019densefusion} by 1.81\% and 1.51\%.  Figure~\ref{fig:YCB_visual} displays several testing visual results, from which we can learn that our method can achieve promising results in some partial occlusion scenarios.

\section{Conclusion and Future Work}
\label{sec:Conclusion}
In this paper, we present a novel 6D pose estimation framework to learn overall point cloud feature representations, aiming to extract more expressive features to achieve accurate 6D pose estimation. 
During the feature processing stage, we consider the point cloud as a special graph structure and finely design a local feature extractor base on graph convolution network, which can effectively excavate the local geometry and topology relationships embedded in the point cloud.
Subsequently, due to the great associative representational capabilities, we apply Transformer to propagate local information in the global scale to achieve the global point cloud information exchange. 
The key ingredient of the proposed model is geometry-aware module in Transformer Encoder. It introduces graph architecture that allows the model to fully exploit the geometry information contained in the local neighborhood of the point cloud. Furthermore, it plays the role of the inductive bias in the proposed framework, which can form an effective constraint for point cloud learning and assist the model to select a more appropriate model to predict 6D pose.
More essentially, the geometry-aware module enable geometry and topology relations provide a guidance for exchange and sharing of global information.
Ablation studies verify the effectiveness of graph convolution network block and geometry-aware module. Our method utilizes point cloud and achieves results comparable to the state-of-the-art RGBD-based methods on three benchmark datasets, proving that proposed approach is productive.

In the future, we will consider exploring two aspects. First, we will improve the feature extract process and attempt to extract similar features from sparse point clouds of same category objects to achieve category-level pose estimation.
Second, we expect the model to have the capability of few-shot or zero-shot pose estimation. Therefore, we consider introducing large-scale datasets containing multiple classes of objects to pre-train our model, which is essential to further improve the network's performance and generalization ability.

\printcredits

\bibliographystyle{cas-model2-names}

\bibliography{cas-refs}

\end{document}